\documentclass[conference]{IEEEtran}
\IEEEoverridecommandlockouts

\usepackage{cite}
\usepackage{amsmath,amssymb,amsfonts}
\usepackage{algorithmic}
\usepackage{graphicx}
\usepackage{subcaption}
\usepackage{textcomp}
\usepackage{xcolor}
\def\BibTeX{{\rm B\kern-.05em{\sc i\kern-.025em b}\kern-.08em
    T\kern-.1667em\lower.7ex\hbox{E}\kern-.125emX}}
\begin{document}

\title{Mixture GAN For Modulation Classification Resiliency Against Adversarial Attacks}\date{}

\author{\IEEEauthorblockN{
		\normalsize{Eyad Shtaiwi}\IEEEauthorrefmark{1},
		\normalsize{Ahmed El Ouadrhiri}\IEEEauthorrefmark{2},
		\normalsize{Majid Moradikia }\IEEEauthorrefmark{2},
			\normalsize{Salma Sultana }\IEEEauthorrefmark{2},
	\normalsize{Ahmed Abdelhadi}\IEEEauthorrefmark{2},
		\normalsize{and Zhu Han}\IEEEauthorrefmark{1},\\}
	\IEEEauthorblockA{
		\IEEEauthorrefmark{1}\normalsize{Electrical and Computer Engineering Department, University of Houston, Houston, TX, USA.}\\
		\IEEEauthorrefmark{2}\normalsize{Department of Engineering Technology, University of Houston, Houston, TX, USA.}\\
	}
	\vspace{-1cm}
}
 \maketitle
 \vspace{0.2cm}

\begin{abstract} 
Automatic modulation classification (AMC) using the Deep Neural Network (DNN) approach outperforms the traditional classification techniques, even in the presence of challenging wireless channel environments. However, the adversarial attacks cause the loss of accuracy for the DNN-based AMC by injecting a well-designed perturbation to the wireless channels. In this paper, we propose a novel generative adversarial network (GAN)-based countermeasure approach to safeguard the DNN-based AMC systems against adversarial attack examples. GAN-based aims to eliminate the adversarial attack examples before feeding to the DNN-based classifier. Specifically, we have shown the resiliency of our proposed defense GAN against the Fast-Gradient Sign method (FGSM) algorithm as one of the most potent kinds of attack algorithms to craft the perturbed signals. The existing defense-GAN has been designed for image classification and does not work in our case where the above-mentioned communication system is considered. Thus, our proposed countermeasure approach deploys GANs with a mixture of generators to overcome the mode collapsing problem in a typical GAN facing radio signal classification problem. Simulation results show the effectiveness of our proposed defense GAN so that it could enhance the accuracy of the DNN-based AMC under adversarial attacks to 81\%, approximately. \vspace{0.2cm}
\end{abstract}


\section{Introduction} 
Automatic modulation classification (AMC) is an approach that is used to automatically recognize the modulation classes of the received signals without any prior knowledge. AMC plays an important role in many military and civilian communication systems \cite{introduction1,introduction2,introduction3}. Generally, the AMC approaches fall into one of two categories \cite{introduction4}. The first category is based on the maximum likelihood that relies on the deployed statistical model, and their performance depends on the accuracy of the considered system model which make them susceptible to the model mismatch \cite{introduction5,introduction6}. The second category deploys the feature-based learning approach that collects the features from the received data samples and uses a classifier at the receiver to determine the modulation classes \cite{introduction7}. Particularly, the authors have used the convolutional neural networks (CNNs) for the classical task of AMC. \vspace{0.1cm} \par
Despite their great results, deep neural networks (DNNs) have been shown to be vulnerable against adversarial attacks \cite{intro_attack1,intro_attack2}. Adversarial examples cause misclassifications of DNN-based classifier by incorporating the original input with a small and carefully designed perturbations. The authors in \cite{intro_attack2} showed that the DNN-based AMC systems are susceptible to adversarial attacks. Moreover, targeted fast gradient method (FGM)-based adversarial examples \cite{book1FGM} has been generated to enforce the misclassification at the receiver to the target label. However, the aforementioned adversarial examples are affected by the channel effects. To mitigate this issue, in \cite{attack_robust}, the authors have proposed some approaches to make the adversarial attack robust against realistic channel effects.  \par
\vspace{0.1cm} Recently, given the power of Generative Adversarial Networks (GAN) \cite{gan1}, the idea of defense-GAN is proposed in \cite{samangouei2018defense} to safeguard the DNN against adversarial attacks. The authors have shown the effectiveness of defense-GAN against both black and white-box attacks. However, the typical DNN classifier considered in \cite{samangouei2018defense}  aimed to classify the images in the presence of an adversary. Therefore, no communication system was considered, and thus their proposed defense does not work for the radio signal classification scenario, as we intended here. Moreover, training of the GAN for the radio signals is challenging as it is easily trapped into the mode collapsing issue where the generator only concentrates to generate samples lying on a few classes instead of the whole data space \cite{goodfellow2014generative}. \vspace{0.1cm} \par  To deal with this concern, motivated by the idea of Mixture Generative Adversarial Nets (MGAN) proposed in \cite{mgan}, we develop a novel approach where our GAN is equipped with a mixture of generators. More explicitly we deploy multiple generators, instead of using a single generator as in the typical GAN. Our simulation results validate that our proposed defense-GAN could achieve an acceptable protection, i.e., the promising accuracy of approximately 81\%, against one effective kind of attacks Fast Gradient Sign Method (FGSM). 
\vspace{0.1cm}
\par The rest of this paper is organized as follows. We describe the system in Section \ref{sec:system}. Section \ref{sec:classifier} details the DNN classier. Section \ref{sec:attack} presents the FGSM-based adversarial attack. Then, we introduce the proposed MGAN in Section \ref{sec:proposed}. Section \ref{results}  presents the simulation results.  Finally, Section \ref{conc} concludes the paper.

\section{System Model}
 \label{sec:system}
\begin{figure}[th!]
	\centering
	\includegraphics[scale=0.2]{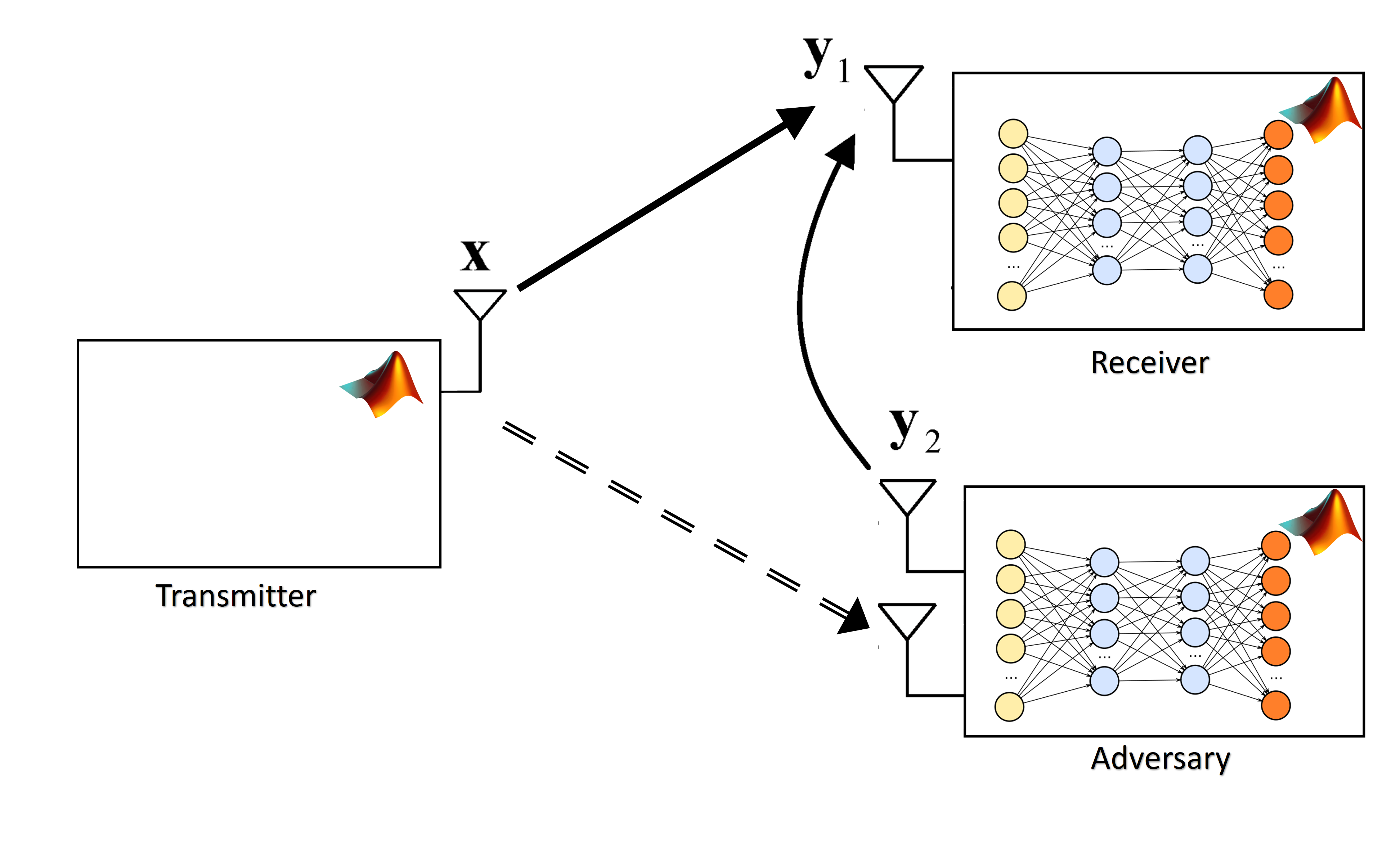}
	\caption{System Model}
	\label{systemmodel}
\end{figure}
As shown in Fig. \ref{systemmodel}, we consider a wireless communication system that consists of three nodes including  transmitter (Tx),  receiver (Rx), and  an adversary (Ad). Next, we describe the working principle of each node as follows. 
\begin{figure*}[ht]
	\centering
	\includegraphics[scale=.61]{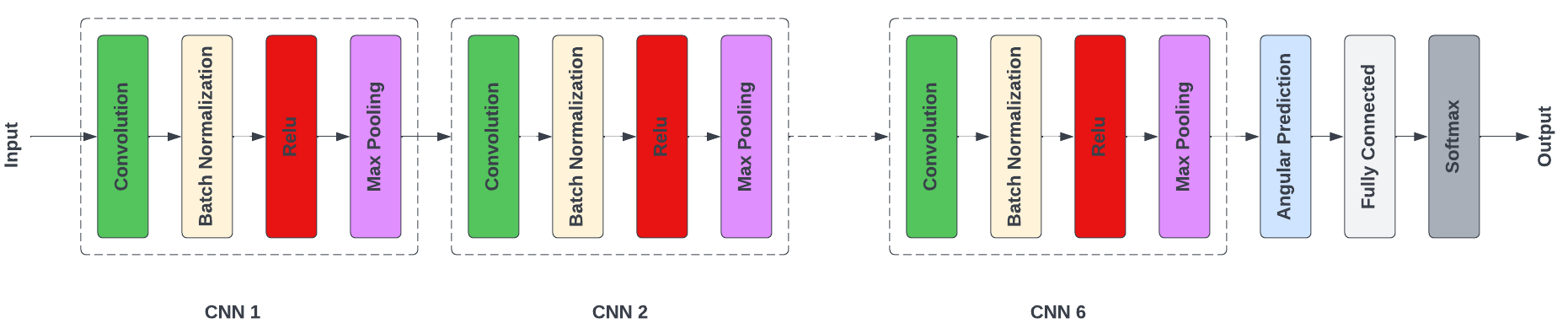}
	\caption{The CNN architecture.}
	\label{cnn}
\end{figure*}
 
The transmitted modulated signal over the wireless channel is an $L$-dimensional vector denoted by  $\mathbf{x} = \left[x[0], x[1], \ldots, x[L] \right]^T \in \mathcal{C}^L$. The wireless channel interacts with the modulated signal by introducing carrier frequency and phase shift offsets, sample rate offset (SRO), selective fading, and additive white Gaussian noise (AWGN). Let $ h[k]$ be the channel's impulse response including the aforementioned radio imperfection at sample time $k$, and then the received signal is expressed as   
\begin{equation}
    r[k] = x[k]\ast h[k] + n[k], 
\end{equation}
\noindent where $n[k]$ is the complex zero mean AWGN which is modeled as $\mathcal{C N}\left(0,\sigma_n^2\right)$, where $\sigma_n^2$ is the noise power. It is worth mentioning, the received complex signal is represented using a $2$-dimensional reals. Equivalently,  $\mathbf{r} = \left[\mathbf{r}_{\rm I} \; \mathbf{r}_{\rm Q}\right] \in \mathbb{R}^{2 \times L}$, where $\mathbf{r}_{\rm I}$ and $ \mathbf{r}_{\rm Q}$ represent 
the in-phase (I) and quadrature (Q) components of $\mathbf{r}$, respectively. In order to minimize the receiver sensitivity to the inter-symbol interference (ISI) and achieve a band-limited  modulated signal, we low-pass filter the I/Q samples through a pulse shaping filter called a  raised cosine filter. The time-domain impulse response of the shaping filter is given by 
\begin{equation}
    p[k] = \frac{\sin{\left(\frac{\pi k}{\tau}\right)} }{{\left(\frac{\pi k}{\tau}\right)}} \frac{\cos{\left(\frac{\pi\rho k}{\tau}\right)}}{1-\left(\frac{2 \pi\rho k}{\tau}\right)^2},
\end{equation}

where $\tau$ and $\rho$ denote the pulse period and the roll-off factor which determines the modulated signal bandwidth, respectively.   
\section{Classifier Architecture}
\label{sec:classifier}
 Let $\mathcal{X}$ denote the set of modulated signals where $\mathcal{X}  \subset \mathbb{R}^{2 \times L}$. Each modulated signal, i.e., $\mathbf{x} \in \mathcal{X}$, belongs to of one of the modulation classes $C$. Therefore, the  ML-based classifier which is denoted by $f\left(:,\mathbf{\theta}\right)$, where maps each frame of the modulated signals into $C$, where $\mathbf{\theta}$ represents the model parameters. In other words, $f\left(:,\mathbf{\theta}_0\right): \mathbb{R}^{L\times 2} \rightarrow \mathbb{R}^{C}$. After aggregation of the data in $\mathcal{X}$, the trained classifier assigns a label $\hat{C}\left(\mathbf{x};\mathbf{\theta}_0\right)$ for each input $\mathbf{x}$ where $\hat{C}\left(\mathbf{x};\mathbf{\theta}_0\right)$ is given by 
\begin{equation}
   \hat{C}\left(\mathbf{x};\mathbf{\theta}_0\right)= \operatorname*{arg\,max}_n \; \; g_n(x,\mathbf{\mathbf{\theta}_0}),
\end{equation}
where $f_n(x,\mathbf{\mathbf{\theta}_0})$ denotes the classification probability that the signal $\mathbf{x}$ belongs to the $n^{\rm th}$ class for $ n = 1, 2, \ldots, C$. In this paper, we consider  $f\left(:,\mathbf{\theta}_0\right)$ for the CNN architecture. 
\par The adopted CNN-classifier consists of multiple convolutional layers, where each layer is able to extract an underlying spatially correlated information from its input  without need to  complex a priori constraint.  
 
 	\begin{figure}[th!] 
	\centering
\includegraphics[scale=0.3]{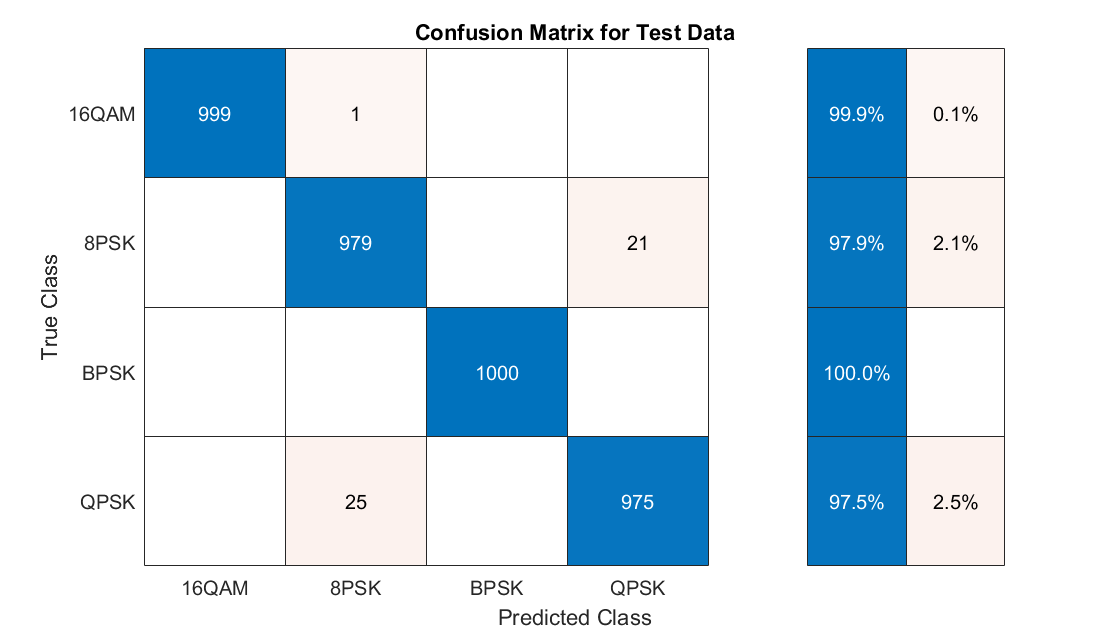}
\caption{Confusion Matrix for the DNN-based AMC.}
\label{a}
	\end{figure}
Fig. \ref{cnn} depicts the CNN-based classifier architecture that consists of six convolution layers, one fully connected layer, and a softmax layer. Each convolutional layer has followed by a batch normalization layer, rectified linear unit (ReLU) activation layer, and a max-pooling layer to form CNN blocks, i.e., CNN1, $\cdots$, CNN6.  In the last convolution layer, the max-pooling layer is replaced with an average pooling layer. The output layer has softmax activation, as well. The $j^{th}$ convolutional layer produces output, i.e., feature map, is given by
	\begin{equation}
		 L_{j}^{(l)} (m,n)=\sigma \big (O_{j}^{(l)} (m,n)\big), 
	\end{equation}

\noindent where $\sigma(\cdot)$ and $O_{j}^{(l)} (m,n)$ represent the activation function, and the weighted sum output of the previous convolutional layer which is computed by 

\begin{align}
	 O_{j}^{(l)} (m,n)=\sum_{i=1}^{M^{(l)}}{\sum_{u,v=0}^{S-1} {{W}_{ji}^{(l)} } } (u,v) L_{i}^{(l-1)} (m-u,n-v)+\!b_{j}^{(l)}, \notag \\ {}
 \end{align}
\noindent where $M$ and $S$ represent the size of the kernel filter, while $b$ and ${W}$ are the bias and the weights parameters that are optimized using the back-propagation (BP) method. Therefore, the updated weights and bias formulas of the $l^{th}$ convolutional layer are 
\begin{align} {W}_{ji}^{(l)}=&{W}_{ji}^{(l)} +\alpha \cdot \frac {\partial L}{\partial {W}_{ji}^{(l)} }, \label{weight}  \\ b_{j}^{(l)}=&b_{j}^{(l)} +\alpha \cdot \frac {\partial L}{\partial b_{j}^{(l)} }.
\label{bias}
\end{align}
Consequently, the derivatives of (\ref{weight}) and (\ref{bias}) are computed using the chain rule as    
\begin{align} \frac {\partial L}{\partial W_{ji}^{(l)} }=&\sum \limits _{m,n} {\delta _{j}^{(l)} } (m,n)\cdot L_{j}^{(L-1)} (m-u,y-v), \\ \frac {\partial L}{\partial b_{j}^{(l)} }=&\sum \limits _{m,n} {\delta _{j}^{(l)} } (m,n),
 \end{align}
where $\delta _{j}^{(l)} $ represents the error map and is given as 
\begin{equation} \delta _{j}^{(l)} =\sum \limits _{j} {\sum \limits _{u,v=0}^{S-1} {\mathcal{W}_{ji}^{(l+1)} } } (u,v)\cdot \delta _{j}^{(l+1)} (m+u,n+v). \end{equation}

After training the model with RML2016.10a dataset \cite{dataset}, we tested the trained model for 4 modulation types, namely, BPSK, QPSK, 8PSK, and 16QAM. The resultant confusion matrix is shown in Fig. \ref{a}, has been obtained using $10^4$ frames per modulation. \par 
It can be seen that the accuracy approximately reached 99\%, and the CNN classifier is barely confused between 8PSK and QPSK. 
%
%
%
%
%
%
 ~\\
\section{Adversarial Attack for DNN-based Classifier Model}
\label{sec:attack}
 The DNN-based AMC systems are usually vulnerable to various security attacks. For instance, different adversarial attacks cause either a misclassification or loss of the accuracy of the DNN-based modulation classifiers. Generally, the attacker in different wireless applications aims to misclassify the CNN-based classifier by transmitting a well-designed perturbation over the wireless channel \cite{intro_attack2}.  \par The current adversarial attacks on the DNN-based AMC systems adopt the white-box attack.  This type of attack setting assumes that adversarial node fully accesses to the the internal information of DNN platforms, such as trained model of $f\left(:,\mathbf{\theta}\right)$ and its parameters and architecture. However, this assumption is almost impossible in real case scenarios as the model owners do not disclose any information about their models. Hence, we develop an black-box adversarial attack against the adopted DNN-classifier. \par The adversarial node affects the the classification decision process for the legitimate receiver DNN-classifer, i.e., $f\left(:,\mathbf{\theta}\right)$, by  injecting adversarial interference into transmitted signals. The perturbed signals can be modeled as \cite{attack_robust} 
\begin{equation}
    \mathbf{x}_{\rm b} = \mathbf{x} + \boldsymbol{\varrho}, 
\end{equation}
where $\varrho \in \mathbb{R}^{L \times 2}$ denotes the additive adversarial interference which is computed as follow 
\begin{align} &\operatorname*{argmin}_{\boldsymbol{\varrho}} \vert \vert \boldsymbol{\varrho} \vert \vert_{\infty}\\ &\text { s.t. } \\ &\hat{\mathcal{C}}(\mathbf{x}\; ;\mathbf{\theta}_0), \neq \hat{\mathcal{C}}(\mathbf{x}+\boldsymbol{\varrho}\; ;\mathbf{\theta}_0) \\ & \mathbf{x}_b+\boldsymbol{\varrho} \in \mathcal{X},   \label{attack1}\end{align}
 where $ \vert \vert \cdot  \vert \vert_{\infty}$ denotes the $l_{\infty}$ norm. It is noteworthy that solving (\ref{attack1}) is difficult and not global optimal. Hence, the generated perturbation, i.e., $\boldsymbol{\varrho}$ is not unique and there are sub-optimal approaches to compute $\boldsymbol{\varrho}$. The most common approach among these sub-optimal methods is the FGSM \cite{desc4}. Specifically, the FGSM crafted the perturbation signal as follows 
  \begin{equation}
\tilde{\mathbf{x}}_{b} = \mathbf{x}+\eta \cdot \operatorname{sgn}\left(\nabla_{x} J\left(\mathbf{x}, y, \boldsymbol{\theta}_0\right)\right),
 \end{equation}
 where $\operatorname{sgn}\left( \cdot \right)$, $\nabla_{x} J\left(\mathbf{x}, y, \boldsymbol{\theta}_0\right)$ denote the sign operation, and the gradient of the model loss function which is function of the input sample, $\boldsymbol{x}$, the model parameters $\boldsymbol{\theta}_0$, and the label vector $\boldsymbol{y} = \{0, 1 \}^C$.  The FSGM incorporates a small perturbed signal, $\boldsymbol{\eta}$,  into each feature of the input sample in the direction of the sign of the classifier's cost function, i.e., $J\left(\mathbf{x}, y, \boldsymbol{\theta}_0\right)$. 
 
\vspace{0.2cm} It is worth mentioning that there are two variants of the FGSM attacks, namely, the targeted FGSM and untargated FGSM. In the first type, the ad node aims to  generate the perturbation that causes the DNN-based classifier to have a predetermined  misclassification, e.g., the DNN-based AMC classifies BPSK modulation
as 8PSK modulation. While, the adversarial node does not select any specific misclassification when computing
the perturbation signal. Based on the output modulation class of the CNN classifier and the attack type, the ad node apply the FGSM method to compute the needed perturbation to confuse the Rx node  classifier.   
Since the ad node does not have full access to the DNN model, the overall accuracy of the substitute classifier is almost 92.5\%.
\section{GAN-based Defense Proposed Approach}
\label{sec:proposed}
In this section, we introduce the countermeasure approach to mitigate the effect of the adversarial interference  and improve the robustness of the legitimate DNN-based classifier. Specifically,  we propose a GAN model as a defensive approach against the attacker. A GAN is a type of neural network framework for the generative modeling approach which uses an existing distribution of samples from a dataset to generate new instances that follow the same distribution of the training dataset \cite{GAN_1}. Moreover, A GAN is considered as a generative model that is trained using two neural network models: $1)$ The generative network model, also known as generator $G$, which aims to learn to generate new plausible samples similar to the training dataset. $2)$ The discriminative network model, also known as discriminator $D$, which is trained to differentiate between the generated samples by the generator and real samples of the training dataset. The generator and the discriminator are set up in a game during the training; the generator aims to fool the discriminator while the discriminator aims to distinguish between real and generated samples. The objective of the generator and the discriminator is to minimize the following min-max loss \cite{GAN_1}
\begin{equation} 
\begin{split}
\min_{G} \max_{D} V (D, G) & = \mathbb{E}_{d \sim  p_{data}(d)}[\log D(d)] + \\
 &  \mathbb{E}_{h \sim p_{h}(h)} [\log (1 - D(G(h)))].
\end{split}
\end{equation}
\par In this paper, we deploy the GAN model to generate plausible samples similar to the received frames. Then, we compare the generated frames with the perturbed received signal, i.e., $\boldsymbol{x}_b$, to determine the true class of the modulated signal, $\mathbf{x}$. Since we are dealing with multiple modulation types, we faced the mode collapse problem, where the generator fails to generate samples of all modulation types. For example, in our experiments, sometimes the generator only generates frame samples that belong to modulation types BPSK, QPSK, and 8PSK. To overcome this problem, we propose to use Mixture GAN (MGAN) \cite{mgan} that allows to avoid the mode collapse problem and increases the defense GAN accuracy. 
%
\vspace{0.2cm}
\par Therefore, we train a GAN model for each modulation type instead of training one GAN for multiple labels. Consequently, in our case, we will have four generative network models; a generator for each modulation type. The proposed approach using four generators does not add any complexity to the system considering the offline training. However, this approach increases the model accuracy and robustness against the ad node attacks. 
After obtaining the multiple generators, the actual modulation class of the received signal in the presence of adversarial attacks can be found as follows  

\begin{equation}
	\label{eq:minimization}
	h^* = \arg \min_{i} \arg \min_{h} ||G_i(h) - r||_2^2 
\end{equation}

\noindent Where $r$ is the received signal after the ad node injecting the perturbation signal, and $i = 1, ..., C$. Afterward, $h^*$ is fed to the corresponding generator $G_i$ to generate a signal close to the received perturbed signal. The optimization problem in (\ref{eq:minimization}) is a highly non-convex problem and has no optimal global solution. Therefore, we adopted  the  gradient descent (GD) method to find the sub-optimal solution \cite{samangouei2018defense}. The developed GD method is equivalent to computing a $q$ gradients of the loss function using $\zeta$ different random initialization of $h$ for each generator as 
  
	\begin{align}
   h^{\zeta}_{q} =& h^{\zeta}_{q - 1} + \eta_{q - 1} \nabla_h \mathcal{L} (r, h)|_{h_{q}= h_{q - 1}}, \forall q = 1, 2, \ldots,  
	\end{align}
 where  $\mathcal{L}(r, h) = ||G(h) - r||_2^2$. 




\begin{figure}[ht]
	\begin{subfigure}{0.5\textwidth}
		\includegraphics[width=\textwidth]{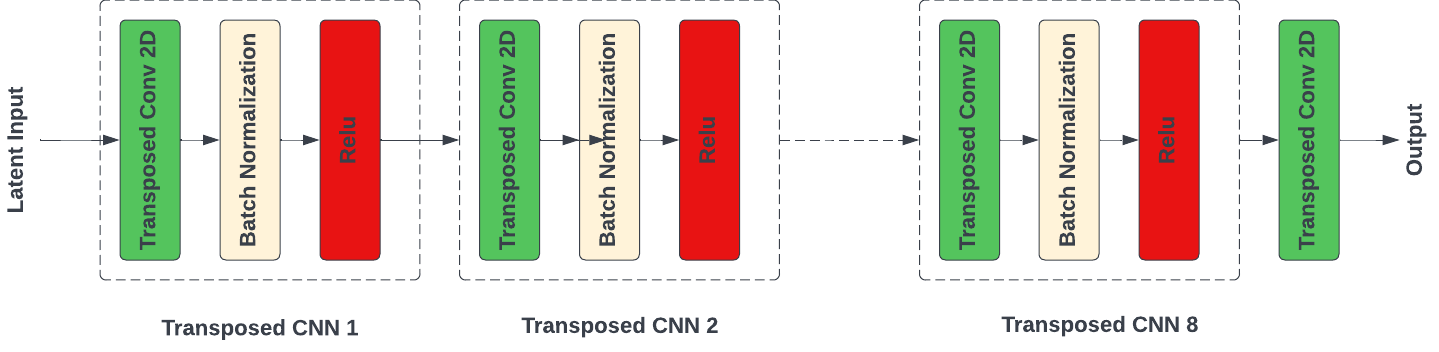}  
		\caption{Generative Network Model Architecture}
		\label{fig: sub_gen}
	\end{subfigure} 
	\begin{subfigure}{0.5\textwidth}
		\includegraphics[width=\textwidth]{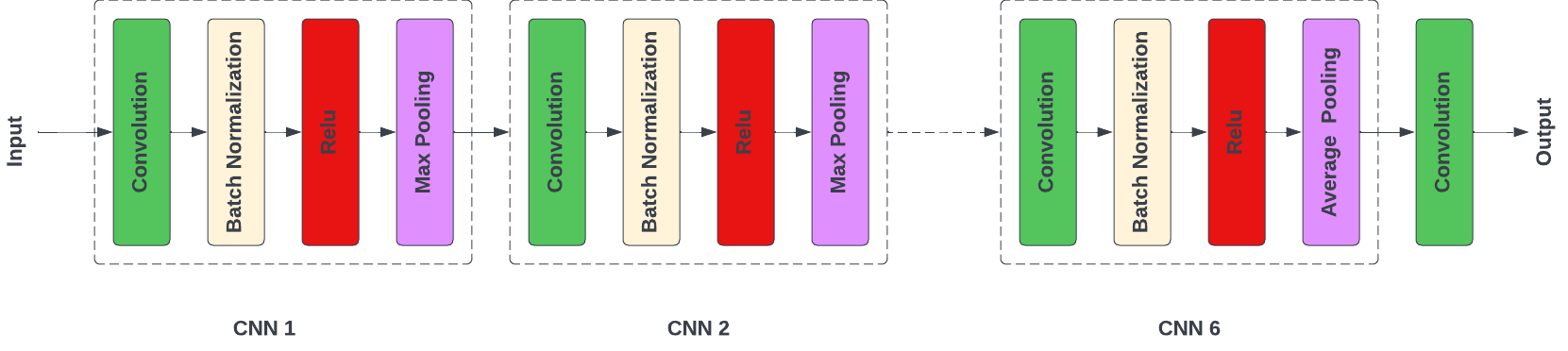}  
		\caption{Discriminative Network Model Architecture}
		\label{fig: sub_disc}
	\end{subfigure}
 
	\caption{GAN model architecture which consists of two neural network models: Generator and Discriminator}
	\label{fig: GAN architecture}
\end{figure}

Fig.\ref{fig: GAN architecture} illustrates the architecture of the GAN developed in this paper, which includes two network models:

\begin{itemize}
\item \textbf{Generative Network Model:} which takes as input a vector of random values, i.e., latent inputs, and generates frames similar to the training dataset. The developed generator consists of eight transposed convolution neural network layers (TCNN) as shown in Fig. \ref{fig: sub_gen}. The vector of latent inputs is shaped to the proper format before feeding it to the first TCNN. Each TCNN is followed by a batch normalization and a Relu activation layer. The last layer is a TCNN that outputs the generated frame of dimension $\mathbf{x}_{gen} \in \mathbb{R}^{L \times 2} \subset \mathcal{X}$.
~\\ 
\item \textbf{Discriminative Network Model:} takes as inputs both  the actual frames, i.e., $\mathbf{x}$, and the generated frames, $\mathbf{x}_{gen}$ by ${\rm G}_{i}$ $\forall i \in {1, \ldots, C}$. The trained discriminators, i.e., $\rm D_{i}$ $\forall i \in {1, \ldots, C}$ differentiate among the real and the generated frames. Fig. \ref{fig: sub_disc} depicts the developed discriminator architecture which consists of six CNN layers, each CNN layer is followed by a batch normalization layer,  a Relu activation function, and a max-pooling layer, similar to the adopted CNN-classifier in Section \ref{sec:classifier}. The sixth CNN layer is followed by an average pooling layer instead of a max-pooling layer. 
\end{itemize}
 


\begin{figure*}[t!] 
    \centering
        \includegraphics[width=0.45\linewidth, height=0.25\textwidth]{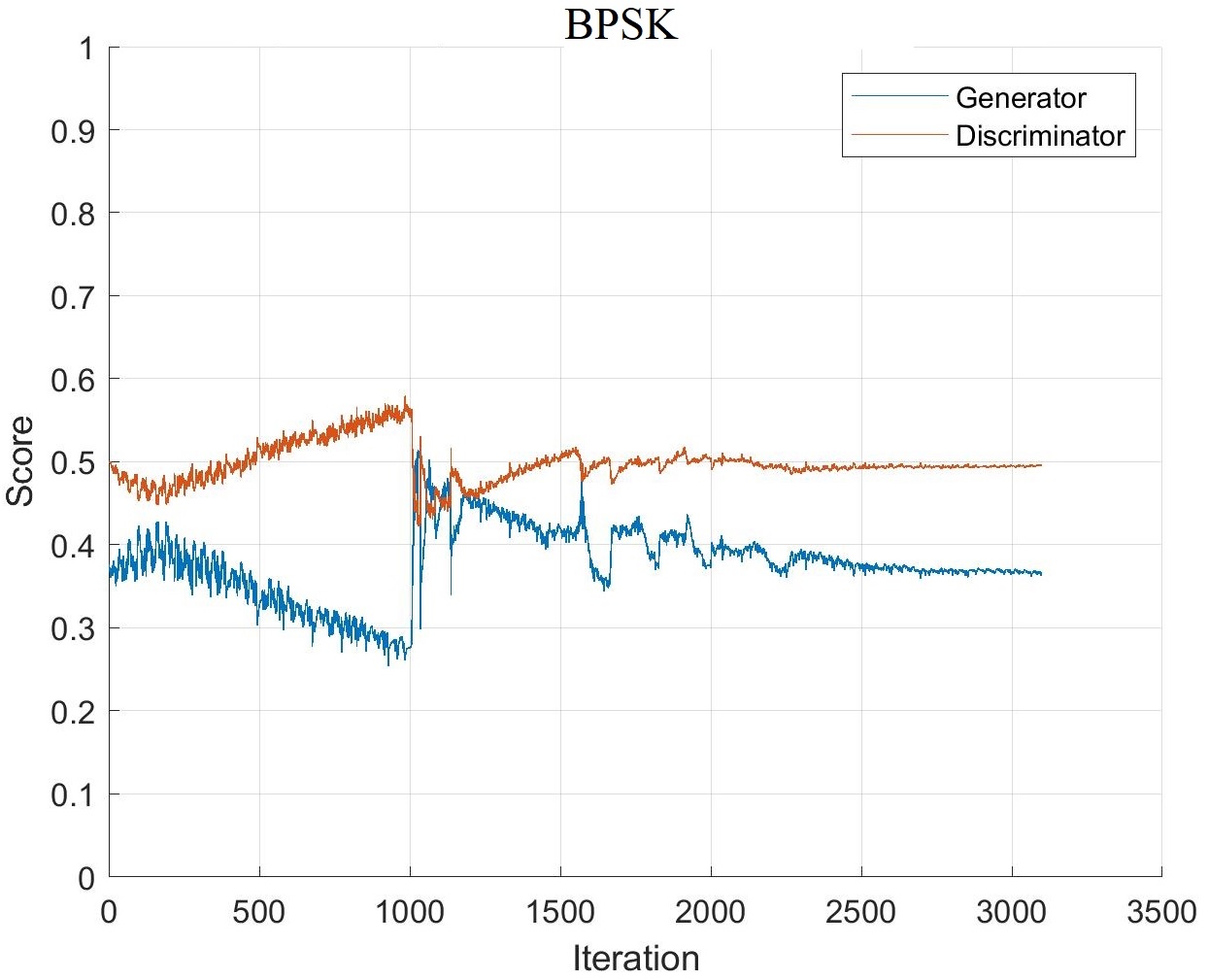}
        \label{fig:gull}
        \includegraphics[width=0.45\linewidth, height=0.25\textwidth]{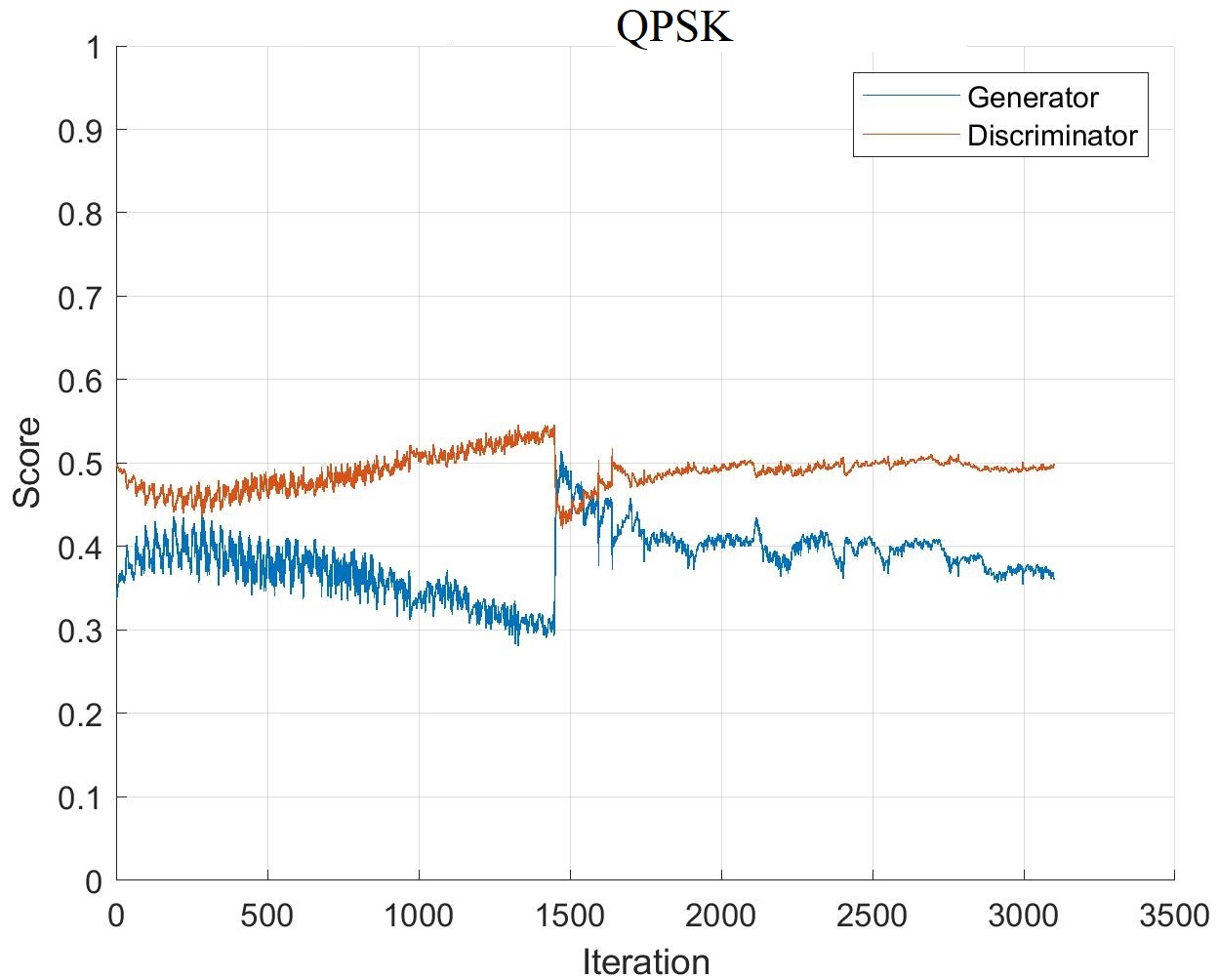}
        \label{fig:tiger}
        \includegraphics[width=0.45\linewidth, height=0.25\textwidth]{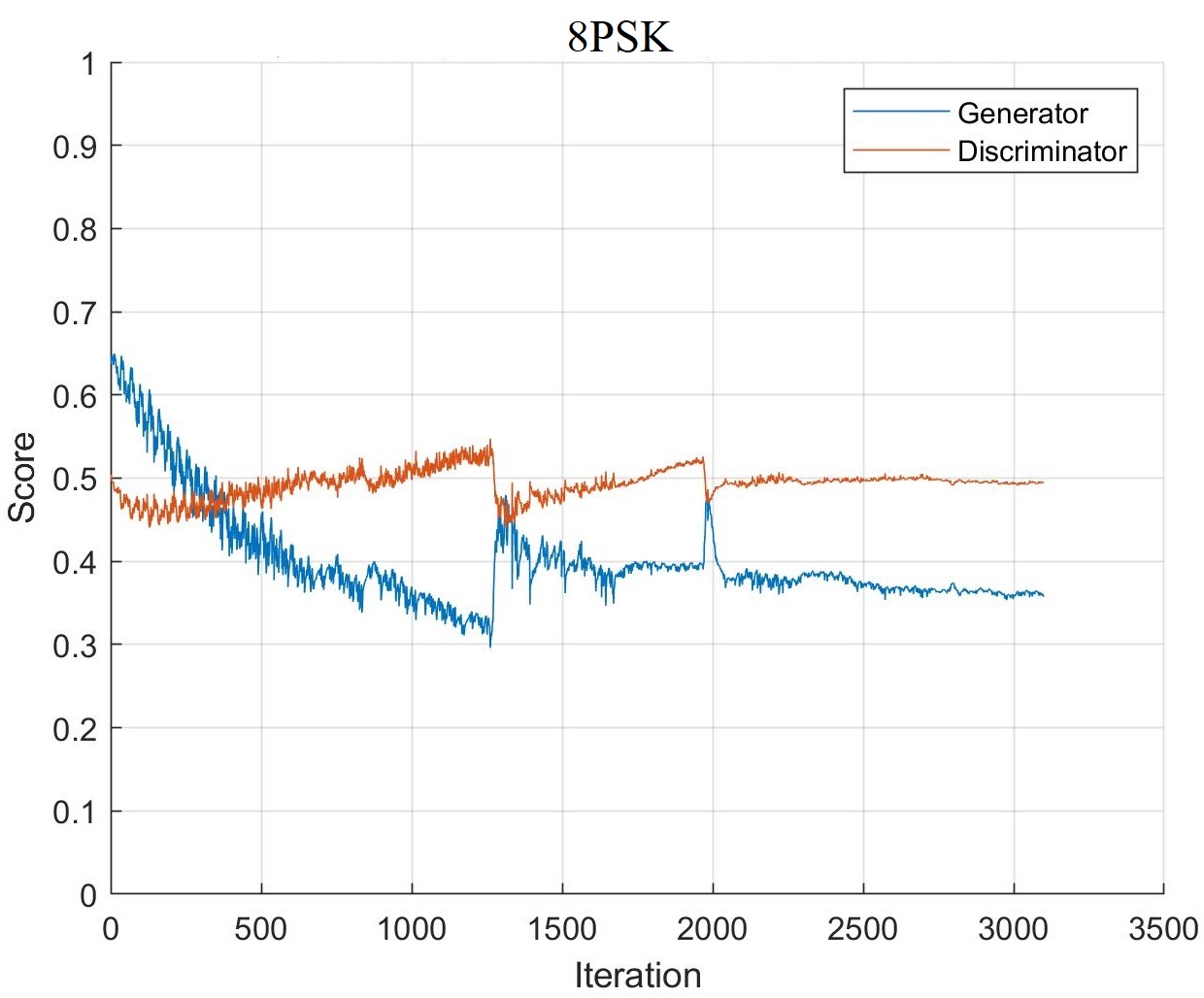}
        \label{fig:gull}
        \includegraphics[width=0.45\linewidth, height=0.25\textwidth]{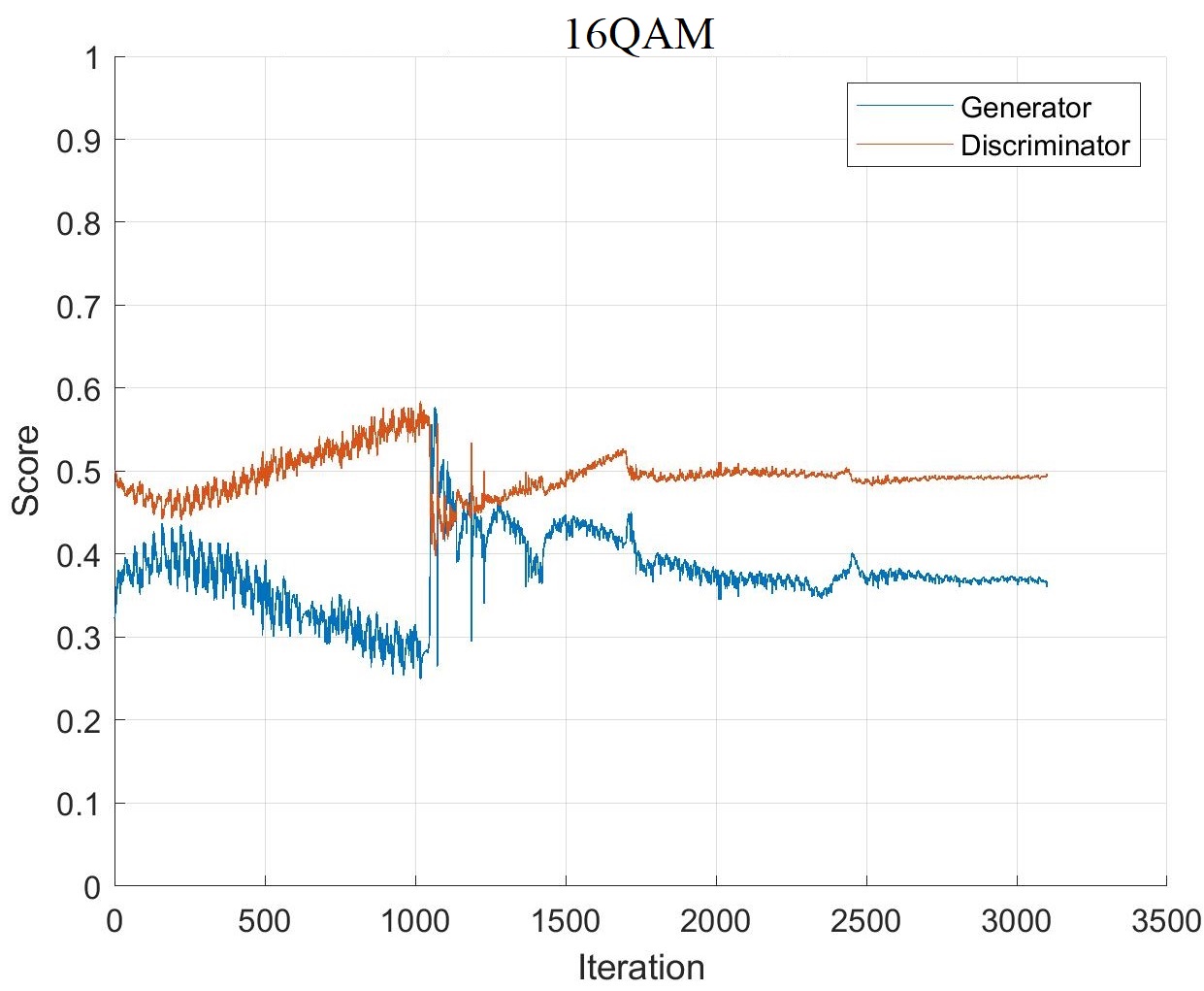}
        \label{fig:tiger}
    \caption{The mixture GAN training for the four modulation types,i.e., BPSK, QPSK, 8PSK, 16QAM.}
    \label{fig: GAN training convergence}
\end{figure*}

\section{Simulation Results}
\label{results}
In this section, we investigate the effectiveness and efficiency of the proposed approach against the adversarial attacks.  Therefore, we study the impact of the presence of the adversarial node on the accuracy of the developed CNN-based classifier in Section \ref{sec:classifier}. Subsequently, we compare the accuracy of the classifier after incorporating the multiple generators with the CNN-classifier. Specifically, we provide the confusion matrices for the classifier for the two cases. 
\vspace{0.2cm}\par In our simulations, we used Matlab platform for training and validating each of the classifier, the attacker, and the mixture GAN on a workstation with Nvidia Titan RTX, $2$ CPUs Intel Xeon Gold $6148$, and $768$GB RAM. The simulation setup parameters are summarized in Table \ref{table11}.  For dataset generation, we consider a Rician multi-path fading channel and introduced channel impairments, such as, carrier frequency and phase offsets. 
	\begin{table}
\begin{center}

			\caption{Simulation Parameters}
					\label{table11}
\begin{tabular}{ | c|c | p{5cm} |}
	
		\hline
		Parameter & Value \\ \hline
 		Carrier Frequency & 915 MHz\\ \hline
		Roll-off factor & 0.7\\ \hline
		Number of frames & $10^4$  \\ \hline
		$L$ & 1024 \\ \hline
		$C$ & 4\\ \hline
		Mini-batch size & 12\\ \hline
		Number of Epochs & 100 \\ \hline
		Optimizer & ADAM \\ \hline
		Learning Rate & 0.02 \\ \hline
		gradient decay factor  & 0.5 \\ \hline

		SNR & $30 {\rm dB}$\\ \hline
		\end{tabular}

\end{center}
\end{table}
 
\subsection{Training of MGAN}
\par In this part, we present the training phase for the Mixture GAN with $4$ generators. In Fig. \ref{fig: GAN training convergence}, we study the convergence performance for both networks, i.e., G's and D's, of each class. It can be observed that the training process succeeded for each modulation class. Specifically, the discriminators reach the optimal score $0.5$ after $1,000$ iterations. On the other hands, the generators converge to $~0.4$ score. Which means that although the discriminators provide the optimal performance, the generators are able to produce modulated signals similar to the actual dataset. Fig. \ref{fig: GAN training convergence} depicts the benefits of deploying multiple generators to overcome the collapsing mode issue of the traditional GAN.
\subsection{Adversarial attack}

\begin{figure}[t!]
	\centering
	\includegraphics[scale= 0.12]{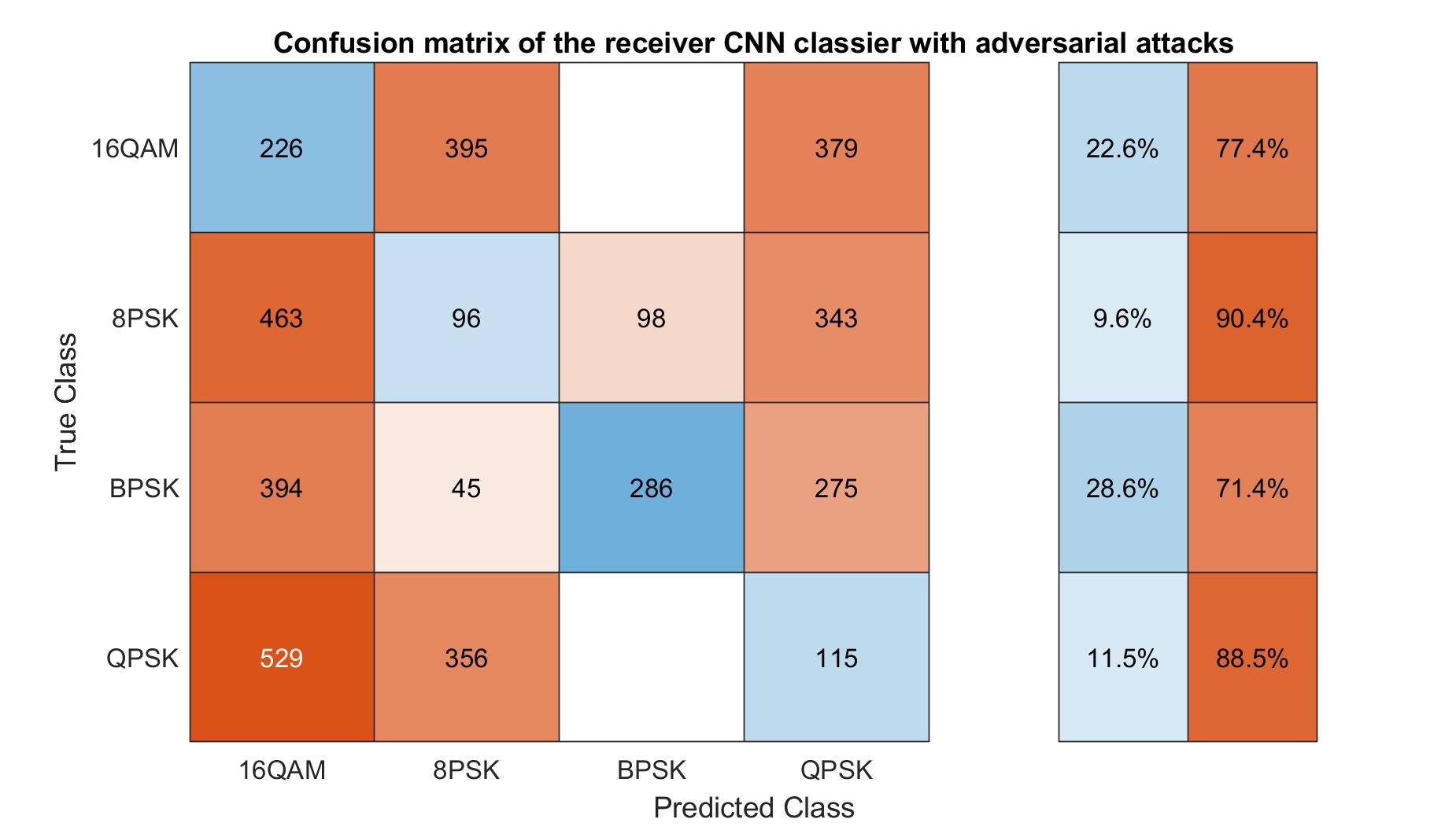}
	\caption{Accuracy of the receiver CNN classifier in the presence of adversarial attacks.}
	\label{fig: Rx CNN classifier with attack}
\end{figure}

 
Fig. \ref{fig: Rx CNN classifier with attack} shows the confusion matrix for the DNN-classifier in the presence of an adversarial attack. In this paper, we only considered the untargeted FGSM variant. Compared with Fig. \ref{a}, it can be noticed that the designed perturbation signals using the FGSM approach reduce the accuracy of the adopted CNN-classifier.It can be shown that the attacker achieves its best performance when the modulated signals are $8$ PSK. Generally, the accuracy performance of the classifier is significantly dropped which proves the effectiveness of the designed attack model.

\subsection{GAN-based countermeasure}

\par
\begin{figure}[t!]
	\centering
	\includegraphics[scale= 0.15]{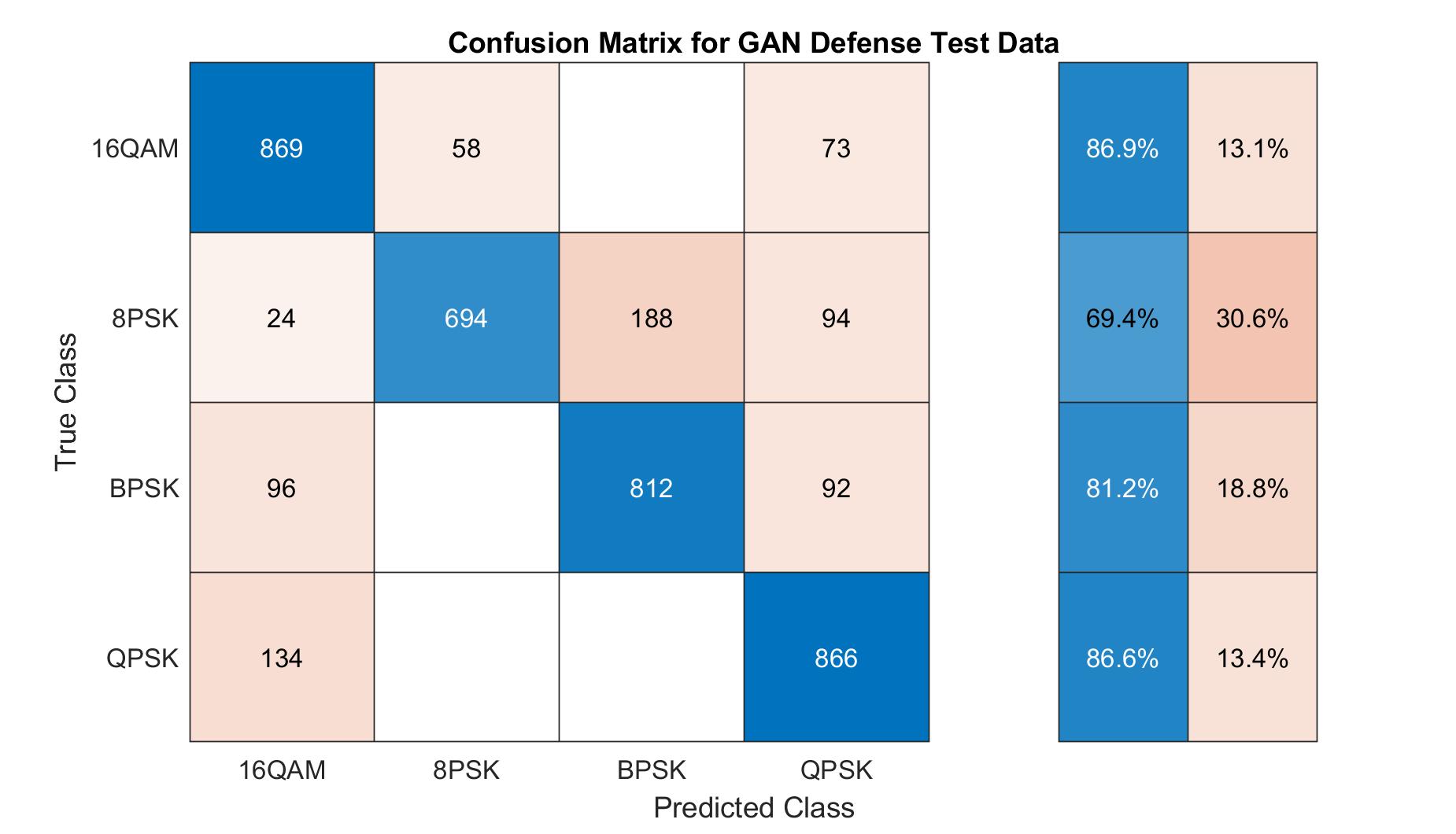}
	\caption{Accuracy of the GAN-based countermeasure.}
	\label{gan_4modulation}
\end{figure}

 Fig. \ref{gan_4modulation} shows the confusion matrix of the classifier in the presence of attack operations after incorporating of MGAN. The average accuracy for the classifier with the proposed defense approach reaches $81 \%$. It can be shown that mixture GAN approach improves  the DNN-based classifer accuracy. Specifically, the accuracy for $8$ PSK case has been improved from $9\%$ to approximately $70\%$. It is worth mentioning, that the error of this confusion matrix includes the error of the GAN model and the CNN-classifier. 


\section{Conclusion}
\label{conc}
 
In this paper, we presented a countermeasure approach against adversarial attacks in AMC systems using mixture GAN. The proposed approach reduces the effect of the adversarial perturbed signals before feeding to the DNN-classifier. To prove the effectiveness of the proposed approach, we alleviated the FGSM approach to craft the perturbation. We utilized the black-box model for the attack node. The adversarial attacks reduce the DNN-classifier accuracy by injecting pre-designed perturbation signals. In addition, we addressed the collapsing mode issue of the traditional GAN by developing multiple generators. The idea of deploying multiple generators helps to capture different modes that exist in the dataset distribution. Finally, through the simulations, we demonstrated the enhancement of accuracy of the CNN-based classifier after incorporating the MGAN method.  


\end{document}